\documentclass[conference]{IEEEtran}
\IEEEoverridecommandlockouts
\usepackage{amsmath,amssymb,amsfonts}
\usepackage{algorithmic}
\usepackage{graphicx}
\usepackage{textcomp}
\usepackage{mathtools}
\usepackage[style=ieee,doi=false,isbn=false,url=false,eprint=false,backend=bibtex]{biblatex}
\usepackage{subcaption}
\usepackage{todonotes}
\usepackage{hyperref}
\usepackage[capitalize]{cleveref}

\let\oldequation\equation
\renewcommand{\equation}{\oldequation\small}

\usepackage{stmaryrd}

\usepackage[mathscr]{eucal}

\usepackage{amsbsy}

\usepackage{bm}

\usepackage{bbding}

\usepackage{array}

\usepackage{tabularx}

\usepackage{ctable}

\usepackage[neveradjust]{paralist}

\usepackage{fancyvrb}

\usepackage{caption}

\usepackage{placeins}

\usepackage{ifpdf}
\ifpdf
\DeclareGraphicsExtensions{.pdf,.png}
\else
\DeclareGraphicsExtensions{.eps}
\fi

\usepackage{ulem}
\usepackage{xcolor}

\newcommand{\email}[1]{\href{mailto:#1}{\nolinkurl{#1}}}

\newcommand{\Sec}[1]{\hyperref[sec:#1]{\S\ref*{sec:#1}}} 
\newcommand{\Section}[1]{\hyperref[sec:#1]{Section~\ref*{sec:#1}}} 
\newcommand{\AppFull}[1]{\hyperref[sec:#1]{Appendix~\ref*{sec:#1}}} 
\newcommand{\Eqn}[1]{\hyperref[eq:#1]{(\ref*{eq:#1})}} 
\newcommand{\Fig}[1]{\hyperref[fig:#1]{Figure~\ref*{fig:#1}}} 
\newcommand{\Tab}[1]{\hyperref[tab:#1]{Table~\ref*{tab:#1}}} 
\newcommand{\Thm}[1]{\hyperref[thm:#1]{Theorem~\ref*{thm:#1}}} 
\newcommand{\Cor}[1]{\hyperref[cor:#1]{Corollary~\ref*{cor:#1}}} 
\newcommand{\Alg}[1]{\hyperref[alg:#1]{Algorithm~\ref*{alg:#1}}} 
\newcommand{\Def}[1]{\hyperref[def:#1]{Definition~\ref*{def:#1}}} 

\newcommand{\Real}{{\mathbb R}}

\definecolor{lightblue}{RGB}{102,217,255}

\newcommand{\Tra}{^{{\sf T}}} 
\newcommand{\V}[1]{{\bm{\mathbf{\MakeLowercase{#1}}}}} 

\newcommand{\M}[1]{{\bm{\mathbf{\MakeUppercase{#1}}}}} 
\newcommand{\MC}[2]{\V{#1}_{#2}} 
\newcommand{\Mn}[2]{\M{#1}^{(#2)}} 
\newcommand{\MnTra}[2]{\M{#1}^{(#2){{\sf T}}}} 

\newcommand{\T}[1]{\boldsymbol{\mathscr{\MakeUppercase{#1}}}} 
\newcommand{\TFS}[2]{\M{#1}_{{#2}}} 

\newcommand{\fnorm}[1]{\left\lVert \, #1 \, \right\rVert_{F}}

\usepackage{optidef}
\usepackage[ruled]{algorithm2e}

\newcommand{\X}{\T{X}}

\renewcommand{\Mn}[2]{\M{#1}_{#2}} 
\renewcommand{\MnTra}[2]{\M{#1}_{#2}^{\sf T}} 

\newcommand{\proxF}[1]{\text{prox}_{#1}}

\newcommand{\prox}[2]{\proxF{#1}\hspace{-0.2em}\left(#2\right)}

\newcommand{\DiagEntries}[1]{\text{Diag}\left( #1 \right)}
\newcommand{\Trace}[1]{\text{Tr}\left( #1 \right)}

\newcommand{\ForAllK}[1]{\left\{ #1 \right\}_{k \leq K}}

\newcommand{\A}{\M{A}}
\newcommand{\Bk}{\Mn{B}{k}}
\newcommand{\BAll}{\ForAllK{\Mn{B}{k}}}

\newcommand{\Dk}{\Mn{D}{k}}
\newcommand{\Pk}{\Mn{P}{k}}

\newcommand{\blueprint}{\M{\Delta_B}}
\newcommand{\B}{\M{B}}

\newcommand{\Aux}[1]{\M{Z}_{#1}}
\newcommand{\AuxV}[1]{\V{z}_{#1}}
\newcommand{\AAux}{\Aux{\A}}
\newcommand{\BkAux}{\Aux{\Bk}}

\newcommand{\BkConstraint}{\M{Y}_{\Bk}}
\newcommand{\BAllConstraint}{\ForAllK{\BkConstraint}}

\newcommand{\DkAux}{\Aux{\Dk}}
\newcommand{\xaux}{\AuxV{\V{x}}}

\newcommand{\DualVar}[1]{\M{\mu}_{#1}}
\newcommand{\ADual}{\DualVar{\A}}
\newcommand{\BkDual}{\DualVar{\BkAux}}

\newcommand{\blueprintkDual}{\DualVar{\blueprint_k}}

\newcommand{\DkDual}{\DualVar{\Dk}}

\newcommand{\Xk}{\Mn{X}{k}}

\newcommand{\ConstraintIndicator}[1]{\iota_{\text{PF2}} \left( #1 \right)}
\newcommand{\ConstraintIndicatorF}{\iota_{\text{PF2}}}

\newcommand{\regF}[1]{g_{#1}}
\newcommand{\reg}[2]{\regF{#1}\left( #2 \right)}
\newcommand{\lossF}[1]{f_{#1}}
\newcommand{\loss}[2]{\lossF{#1}\left( #2 \right)}

\usepackage{layouts}
\def\BibTeX{{\rm B\kern-.05em{\sc i\kern-.025em b}\kern-.08em
    T\kern-.1667em\lower.7ex\hbox{E}\kern-.125emX}}
\begin{document}

\title{PARAFAC2 AO-ADMM: Constraints in all modes}

\author{\IEEEauthorblockN{Marie Roald}
\IEEEauthorblockA{\textit{Simula Metropolitan Center}\\
\textit{for Digital Engineering}\\
\textit{\& Oslo Metropolitan Univ.} \\
Oslo, Norway \\
mariero@simula.no}
\and
\IEEEauthorblockN{Carla Schenker}
\IEEEauthorblockA{\textit{Simula Metropolitan Center}\\
\textit{for Digital Engineering}\\
\textit{\& Oslo Metropolitan Univ.} \\
Oslo, Norway \\
carla@simula.no}
\\
\and
\IEEEauthorblockN{Jeremy E. Cohen}
\IEEEauthorblockA{\textit{University of Rennes} \\
\textit{Inria, CNRS, IRISA}\\
Rennes, France \\
jeremy.cohen@irisa.fr}
\and
\IEEEauthorblockN{Evrim Acar}
\IEEEauthorblockA{\textit{Simula Metropolitan Center}\\
\textit{for Digital Engineering}\\
Oslo, Norway \\
evrim@simula.no}
}

\maketitle

\begin{abstract}
The PARAFAC2 model provides a flexible alternative to the popular CANDECOMP/PARAFAC (CP) model for tensor decompositions. Unlike CP, PARAFAC2 allows factor matrices in one mode (i.e., evolving mode) to change across tensor slices,
which has proven useful for applications in different domains such as chemometrics, and neuroscience. However, the evolving mode of the PARAFAC2 model is traditionally modelled implicitly, which makes it challenging to regularise it. Currently, the only way to apply regularisation on that mode is with a flexible coupling approach, which finds the solution through regularised least-squares subproblems. In this work, we instead propose an alternating direction method of multipliers (ADMM)-based algorithm for fitting PARAFAC2 and widen the possible regularisation penalties to any proximable function. Our numerical experiments demonstrate that the proposed ADMM-based approach for PARAFAC2 can accurately recover the underlying components from simulated data while being both computationally efficient and flexible in terms of imposing constraints.
\end{abstract}


\section{Introduction}
Tensor decompositions, in particular the CANDECOMP/PARAFAC (CP) model \cite{PARAFAC:Ha70,PARAFAC:Ca70}, have successfully extracted meaningful patterns from complex data in many disciplines including chemometrics \cite{Br97} and neuroscience \cite{MoHaHePaAr06,AcBiBiBr07}. However, the CP model has strict assumptions of multilinearity that can be violated in practice. Another tensor model, PARAFAC2 \cite{tensor:parafac2:Ha72}, relaxes the CP model by allowing for evolving factors in one mode. This relaxation also enables decomposing stacks of matrices of varying size.

\begin{figure}[t]
\centering
\includegraphics[width=\linewidth]{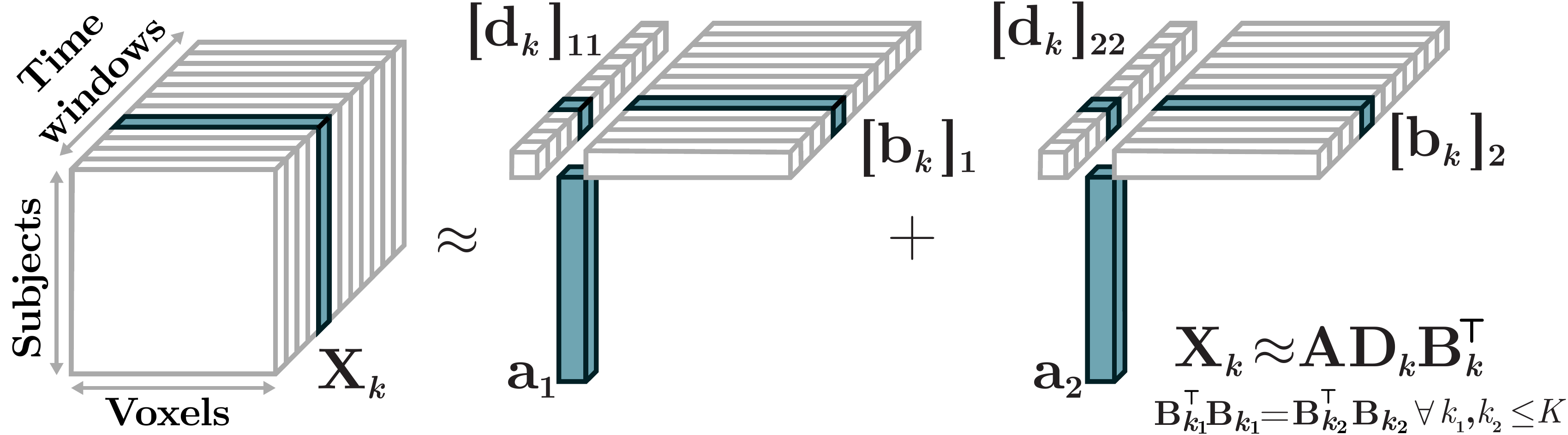}
\caption{Illustration of a two-component PARAFAC2 model for tracing networks in neuroimaging data.}
\label{fig:icassp}
\end{figure}

The ability to describe such evolving or irregular factors has made the PARAFAC2 model a powerful tool. For instance, in chemometrics, PARAFAC2 has been applied to chromatographic data with unaligned elution profiles \cite{parafac2:BrAnKi99}. PARAFAC2 has also been used to analyse unaligned temporal profiles in electronic health records \cite{tensor:smoothness:parafac2:COPA:AfPePaSeHoSe18} and to find information across different languages from a multi-language corpus \cite{ChBaBr07}. Recently, PARAFAC2 has also shown promise for tracing time-evolving patterns of brain connectivity from neuroimaging data (illustrated in \cref{fig:icassp}) \cite{RoShJi19}.

Often, the interpretability of component models, such as CP and PARAFAC2, can be improved through constraints and regularisation. However, evolving components of the PARAFAC2 model are usually computed implicitly \cite{KiTeBr99}. Therefore, it is a challenge to impose constraints or regularisation on these evolving factors. \citeauthor{tensor:smoothness:parafac2:He17} imposed smoothness on these factors by constraining them to follow a low-rank B-spline interpolation \cite{tensor:smoothness:parafac2:He17}. To achieve smoothness, the data tensor is projected onto the linear subspace spanned by the given B-spline interpolation matrix before decomposing with PARAFAC2. However, for this approach to be feasible, knots of the splines must be known a-priori, which may be difficult in practice.

Currently, the only way to regularise the evolving mode of a PARAFAC2 model, without knowing the subspace the components lie in, is with a flexible coupling approach \cite{CoBr18}. This approach relaxes the PARAFAC2 constraint and finds the components by solving regularised least-squares problems. Another notable approach is by Yin \textit{et al.} \cite{YiAfHoChZhSu20} using a regularisation penalty inspired by PARAFAC2 to improve the uniqueness properties of regularised coupled non-negative matrix factorisation for binary data.

In this paper, we propose an alternating optimisation scheme with the alternating direction method of multipliers (AO-ADMM) to fit PARAFAC2 models with regularisation on all modes. The AO-ADMM scheme has recently been introduced to fit tensor models \cite{Huang2016}. In \cite{Huang2016}, Huang \textit{et al.} used AO-ADMM to fit a CP model, and in \cite{ScCoAc20} that was extended to a flexible framework for regularised linearly coupled matrix-tensor factorisations. Afshar \textit{et al.} proposed using AO-ADMM to impose proximable constraints on the non-evolving factor matrices of the PARAFAC2 model \cite{tensor:smoothness:parafac2:COPA:AfPePaSeHoSe18}. Here, we introduce ADMM updates for the evolving mode as well, widening the possible regularisation penalties on this mode to any proximable function. With numerical experiments on simulated data, we show that our approach can accurately recover underlying components while being both flexible in terms of imposing constraints and computationally efficient. 

\section{Tensor decomposition with PARAFAC2}
\emph{Tensors} can be seen as multi-way arrays that generalise the concept of matrices to higher order data \cite{tensordec:KoTaBa09}. As such, a vector is a first-order tensor, a matrix is a second-order tensor, a ``cube'' of numbers is a third-order tensor and so forth. A tensor with more than two modes is often called a \emph{higher-order tensor}. We denote higher-order tensors as \(\T{X}\), matrices as  \(\M{X}\), vectors as \(\V{x}\), and the Frobenius norm of \(\T{X}\) as \(\fnorm{\T{X}}\).

PARAFAC2 can be seen as a relaxed version of the CP model. CP assumes multilinearlity and for a third-order tensor, each frontal slice is modelled as:
\begin{equation}
    \M{X}_k \approx \A \Dk \B\Tra,
\end{equation}
where \(\Dk\) is an \(R \times R\) diagonal matrix. \(R\) is the rank of the decomposition, i.e. the number of components in the model.
Note that each slice, \(\TFS{X}{k} \in \Real^{I \times J}\), has the same \(\A\) and \(\B\) matrices. PARAFAC2, on the other hand, allows each slice to have a different \(\B\) matrix:
\begin{equation}
    \TFS{X}{k} \approx \M{A} \Dk \Bk\Tra, \label{eq:pf2}
\end{equation}
where \(\Bk\)s follow the \emph{PARAFAC2 constraint}, i.e., \(\smash{\MnTra{B}{k_1}\Mn{B}{k_1}} = {\MnTra{B}{k_2}\Mn{B}{k_2}}\) for all \(k_1, k_2 \leq K\), \(\Dk \in \Real^{R \times R}\) is a diagonal matrix.

\section{Optimisation}
\subsection{PARAFAC2 \& ALS} \label{sec:pf2.als}
To solve the unconstrained PARAFAC2 problem, \cite{KiTeBr99} reformulated the model to the following equivalent form:
\begin{equation}
    \TFS{X}{k} \approx \A \Dk \blueprint\Tra \Pk\Tra, \label{eq:pf2.kiers}
\end{equation}
where \(\blueprint\) is a square matrix and \(\Pk\Tra\Pk = \M{I}\). This problem can be solved efficiently using an alternating least squares (ALS) procedure, where the \(\Pk\) updates are performed by solving an orthogonal procrustes problem.

\subsection{PARAFAC2 \& AO-ADMM}
We wish to solve the regularised PARAFAC2 problem
\begin{argmini}|l|
  { \A, \ForAllK{\Bk, \Dk}}
  {\left\{\begin{split}
      & f\left(\A, \BAll, \ForAllK{\Dk} \right) \\
      & + \reg{\A}{\A} + \sum_{k=1}^K \reg{\B}{\Bk} + \reg{\M{D}}{\Dk}
  \end{split}\right\},}
  {}
  {}
  \addConstraint{\Mn{B}{k_1}\Tra \Mn{B}{k_1} = \Mn{B}{k_2}\Tra \Mn{B}{k_2}}{\qquad \forall k_1, k_2\leq K}
\end{argmini}
where \({\small f\left(\A, \BAll, \ForAllK{\Dk} \right)\hspace{-0.2em} = \hspace{-0.2em}\sum_{k=1}^K \fnorm{\A \Dk \Bk\Tra \hspace{-0.2em}- \Xk}^2 }\) is the sum of squared errors (SSE) data fidelity term, and \(\regF{\A}, \regF{\B}, \regF{\M{D}}\) are regularisation functions. However, imposing regularisation is difficult within the traditional ALS algorithm, as it estimates the \(\Bk\) matrices implicitly as the product of orthogonal \(\Pk\) matrices and a \(\blueprint\) matrix.

An alternative to directly solving regularised problems is to use splitting methods. In particular, we use ADMM \cite{BoPaChPeEc11} to solve split problems of the form
\begin{argmini}|l|
  { \V{x}, \xaux}
  {f(\V{x}) + g(\xaux)}
  {}
  {}
  \addConstraint{\V{x}}{= \xaux}
\end{argmini}
Here, \(f\) and \(g\) represent the data-fidelity term and regularisation penalty, respectively.

To use ADMM, we require that the scaled proximal operator \cite{PaBo14} is computationally cheap to evaluate for both the data-fidelity term, \(f: \Real^n \to \Real \), and the regulariser, \(g: \Real^n \to \Real \). For a proper convex lower-semicontinuous function \(h\), the scaled proximal operator with scale parameter \(\rho\) (see \cref{sec:rho} for automatic selection of \(\rho\)), is given by
\begin{argmini}|l|
  { \V{x} \in \Real^n}
  {h(\V{x}) + \frac{\rho}{2}\fnorm{\V{x} - \V{y}}^2.}
  {}
  {\prox{\frac{h}{\rho}}{\V{y}}=}
\end{argmini}
Thus, to apply ADMM to the PARAFAC2 decomposition, we need a natural splitting scheme where all proximal operators are easily evaluated. Such a scheme is known for the static modes of PARAFAC2 \cite{tensor:smoothness:parafac2:COPA:AfPePaSeHoSe18}. However, no ADMM splitting scheme has been presented for the evolving mode yet.

\subsection{ADMM for the B mode}
To specify an ADMM scheme for the problem
\begin{argmini}|l|
  { \BAll}
  {\sum_{k=1}^K \loss{\Bk}{\Bk}+ \reg{\B}{\Bk},}
  {}
  {}
  \addConstraint{\Mn{B}{k_1}\Tra \Mn{B}{k_1} = \Mn{B}{k_2}\Tra \Mn{B}{k_2}}{\qquad \forall k_1, k_2\leq K}
\end{argmini}
where \(\smash{\small \loss{\Bk}{\Bk} = \fnorm{\A \Dk \Bk\Tra - \Mn{X}{k}}^2}\), we introduce two sets of auxiliary variables, \(\BkAux\) and \(\BkConstraint\), which respectively split the regularisation by \(\regF{\B}\) and the PARAFAC2 constraints, forming the problem:
\begin{argmini}|l|
  {\text{\resizebox{.13\textwidth}{!}{\(\ForAllK{\Bk, \BkAux, \BkConstraint}\)}}}
  {\hspace{-0.7em} \sum_{k=1}^K \left[\loss{\Bk}{\Bk} \hspace{-0.1em}+ \hspace{-0.1em}\reg{\B}{\BkAux}\right] \hspace{-0.1em}+\hspace{-0.1em}  \ConstraintIndicator{\hspace{-0.2em}\BAllConstraint\hspace{-0.2em}}\hspace{-0.2em},}
  {}
  {}
  \addConstraint{\hspace{-0.7em}\Bk = \BkAux, \quad \Bk = \BkConstraint}{\qquad \forall k}
\end{argmini}
where {\small \(\ConstraintIndicator{\BAllConstraint} = 0\)} if \(\BkConstraint\Tra\BkConstraint\) is constant over \(k\) and \(\infty\) otherwise.
This problem can be tentatively solved using the ADMM algorithm specified in Algorithm~\ref{alg:admm.b}.

There are three functions whose proximal operator must be implemented for Algorithm~\ref{alg:admm.b}: the data-fidelity function ({\small\(\loss{\Bk}{\Bk} = \fnorm{\A \Dk \Bk\Tra - \Xk}^2\)}), the regularisation function (\(\regF{\B}\)), and the characteristic function for the set of matrices with constant cross product (\(\ConstraintIndicatorF\)). The proximal operator for the data-fidelty function is the least squares solution
\begin{align}
    \prox{\frac{\lossF{\Bk}}{\rho_{\Bk}}}{\M{M}} = \hspace{-0.2em}\left(\Xk\Tra \A \Dk \hspace{-0.15em}+\hspace{-0.15em} \frac{\rho_{\Bk}}{2}\M{M}\right)
    \hspace{-0.2em}\left( \Dk\A\hspace{-0.15em}\Tra  \A \Dk \hspace{-0.15em}+\hspace{-0.15em} \rho_{\Bk} \M{I} \right)^{-1}\hspace{-0.5em}. \label{eq:B.loss.update}
\end{align}
The proximal operator for the regularisation functions is tailored for different regularisation penalties, but can be efficiently computed for a large family of functions.

Unfortunately, the proximal operator for \(\ConstraintIndicatorF\),
\begin{argmini}|l|
  {\hspace{-0.5em}\BAllConstraint}
  {\hspace{-1.5em}\left\{\begin{split}
    &\ConstraintIndicator{\BAllConstraint}\\
    &\hspace{-0.25em}+\hspace{-0.25em}\sum_{k=1}^K \frac{\rho_{\Bk}}{2}\fnorm{\BkConstraint \hspace{-0.5em}- \M{W}_k}^2
  \end{split}\right\},}
  {\prox{\ConstraintIndicatorF}{\ForAllK{\M{W}_k}\hspace{-0.15em}}=\hspace{-0.5em}}
  {}
\end{argmini}
where \(\ForAllK{\M{W}_k}\) is an arbitrary collection of matrices, is not trivial to compute. Nevertheless, it can be approximated with the method of Kiers \textit{et al.} \cite{KiTeBr99}. If we use this method, setting \(\BkConstraint = \Pk \blueprint\) with \(\Pk\Tra\Pk = \M{I}\), we obtain Algorithm~\ref{alg:constraint.prox} for the proximal operator. In our experiments, we found that one iteration of this algorithm was sufficient.

\begin{algorithm}
\small
\DontPrintSemicolon
\SetAlgoLined
\footnotesize
\KwResult{\(\ForAllK{\Pk}, \blueprint\)}
 \While{convergence criteria are not met}{
  \For{\(k \gets 1\) \KwTo \(K\)}{
    Compute ``economy style'' SVD: \( \left(\Bk + \blueprintkDual\right)\blueprint \Tra = \M{U}^{(k)} \M{\Sigma}^{(k)} {\M{V}^{(k)}}\Tra\) \;
    \( \Pk \gets \M{U}^{(k)}{\M{V}^{(k)}}\Tra\) \;
   }
    \(\blueprint \gets \frac{1}{\sum_{k=1}^K \rho_{\Bk}} \sum_{k=1}^K \rho_{\Bk} \Pk\Tra \left( \Bk + \blueprintkDual \right)\) \;
    
 }
 \caption{\small Approximate projection onto set of collections of matrices with constant cross product}\label{alg:constraint.prox}
\end{algorithm}

\begin{algorithm}
\small
\DontPrintSemicolon
\SetAlgoLined
\footnotesize
\KwResult{\(\ForAllK{\Bk, \BkAux, \BkConstraint=\Pk\blueprint}\)}
  \While{convergence criteria are not met}{
    \For{\(k \gets 1\) \KwTo \(K\)}{
      \( \Bk \xleftarrow{\eqref{eq:B.loss.update}} \prox{\frac{2L}{\rho_{\Bk}}}{\BkAux - \BkDual + \BkConstraint - \blueprintkDual} \)\;
      \( \BkAux \gets \prox{ \frac{\regF{\M{B}}}{\rho_{\Bk}} }{\Bk + \BkDual} \) \;
    }
    \(\BAllConstraint \xleftarrow{\text{Alg.}~\ref{alg:constraint.prox}} \prox{\ConstraintIndicatorF{}}{\ForAllK{\Bk + \blueprintkDual}}\) \;
    \For{\(k \gets 1\) \KwTo \(K\)}{
      \( \BkDual \gets \BkDual + \Bk - \BkAux \) \;
      \( \blueprintkDual \gets \blueprintkDual + \Bk - \BkConstraint \) \;
    }
  }
\caption{ADMM for the B mode}\label{alg:admm.b}
\end{algorithm}

\subsection{ADMM for the A mode}
To update the A-mode, we use ADMM to solve the problem
\begin{argmini}|l|
  {\A}
  {\sum_{k=1}^K \fnorm{\A \Dk \Bk\Tra - \Xk}^2 + \reg{\A}{\A}.}
  {}
  {}
\end{argmini}
This requires us to evaluate both the proximal operator of the data-fidelity term, \({\small\loss{\A}{\A} = \sum_{k=1}^K \fnorm{\A \Dk \Bk - \Xk}^2}\):
\begin{equation}
    \prox{\frac{\lossF{\A}}{\rho_\A}}{\M{M}} = \left( \sum_{k=1}^K \Xk \Mn{\Gamma}{k}  + \frac{\rho_\A}{2}\M{M} \right) \left( \sum_{k=1}^K \Mn{\Gamma}{k} \Tra \Mn{\Gamma}{k} + \frac{\rho_\A}{2} \M{I} \right)^{-1}, \label{eq:A.loss.update}
\end{equation}
with \(\Mn{\Gamma}{k} =  \Bk \Dk \), and the proximal operator of the regularisation function, \(\regF{\A}\). With these operators, we obtain the update steps given in Algorithm~\ref{alg:admm.a}.

\begin{algorithm}
\small
\SetAlgoLined
\DontPrintSemicolon
\KwResult{\(\A, \AAux, \ADual\)}
 \While{convergence criteria are not met}{
    \( \A \xleftarrow{\eqref{eq:A.loss.update}} \prox{\frac{\lossF{\A}}{\rho_\A}}{\AAux - \ADual}\) \;
    \( \AAux \gets \prox{\frac{ \regF{\A} }{ \rho_\A }}{\A + \ADual} \) \;
    \( \ADual \gets \ADual + \A - \AAux \) \;
 }
 \caption{ADMM for the A mode}\label{alg:admm.a}
\end{algorithm}

Within the framework of \cite{ScCoAc20}, this approach can be considered as hard coupling for all matrices, \(\Xk\), through \(\A\), and the \(\Bk\) updates would correspond to discovering the structure of the coupling for the \(\Bk\) matrices. 

\subsection{ADMM for the D mode}
The D-mode components are updated independently, finding diagonal matrices that solve the problem
\begin{argmini}|l|
  {\Dk}
  {\fnorm{\A \Dk \Bk\Tra - \Xk}^2 + \reg{\M{D}}{\Dk}, }
  {}
  {}
\end{argmini}
for each \(k\). The proximal operator for the data-fidelity term, \({\small \loss{\Dk}{\Dk} = \fnorm{\A \Dk \Bk\Tra - \Xk}^2}\) is the minimiser of a quadratic function. The minimiser is formulated using the vector containing the diagonal entries of \(\Dk\):
\begin{equation}
    \prox{\frac{\lossF{\Dk}}{\rho_{\Dk}}}{\V{v}} = \left( \A\Tra \A * \Bk\Tra\Bk + \frac{\rho_{\Dk}}{2} \M{I} \right)^{-1}  \left( \V{\xi} + \frac{\rho_{\Dk}}{2}\V{v} \right), \label{eq:C.loss.update}
\end{equation}
where \(*\) is the Hadamard product and \({\small \V{\xi}=\DiagEntries{\A\Tra \Xk \Bk}}\) is the vector containing the diagonal entries of \({\small \A\Tra \Xk \Bk}\). This results in the update steps given in Algorithm~\ref{alg:admm.d}.

\begin{algorithm}
\small
\SetAlgoLined
\DontPrintSemicolon
\KwResult{\(\Dk, \DkAux, \DkDual\)}
 \While{convergence criteria are not met}{
   \For{\(k \gets 1\) \KwTo \(K\)}{
        \( \Dk \xleftarrow{\eqref{eq:C.loss.update}} \prox{\frac{\lossF{\Dk}}{\rho_{\Dk}}}{\DkAux - \DkDual}\) \;
        \( \DkAux \gets \prox{\frac{ \regF{\M{D}} }{ \rho_{\Dk} }}{\Dk + \DkDual} \) \;
        \( \DkDual \gets \DkDual + \Dk - \DkAux \) \;
    }
 }
 \caption{ADMM for the D mode}\label{alg:admm.d}
\end{algorithm}

\subsection{PARAFAC2 AO-ADMM}
By combining the three update algorithms above, we obtain Algorithm~\ref{alg:ao-admm} to fit regularised PARAFAC2 models to data. To measure convergence of the inner loops, we adapted the stopping criteria in \cite{BoPaChPeEc11} with a maximum of five iterations. Likewise, for the outer loops, we used the stopping criteria from \cite{ScCoAc20} with a maximum of 1000 iterations.

\begin{algorithm}
\small
\SetAlgoLined
\DontPrintSemicolon
\KwResult{\(\A, \ForAllK{\Bk, \Dk}\)}
Initialise \(\A, \AAux, \ADual, \Bk, \BkAux, \BkDual, \blueprint, \Pk, \blueprintkDual, \Dk, \DkAux,\) and \(\DkDual\) \;
 \While{convergence criteria are not met}{
    Update \( \ForAllK{\Bk, \BkAux, \Pk, \BkDual, \blueprintkDual}\) and \(\blueprint \) using Algorithm~\ref{alg:admm.b} \;
    Update \( \A, \AAux \) and \(\ADual\) using Algorithm~\ref{alg:admm.a}\;
    Update \(\ForAllK{ \Dk, \DkAux, \DkDual}\) using Algorithm~\ref{alg:admm.d}\;
 }
 \caption{AO-ADMM for PARAFAC2}\label{alg:ao-admm}
\end{algorithm}

\subsection{Selecting \(\rho\)} \label{sec:rho}
For efficient ADMM updates, we need suitable \(\rho\)-parameters. In this work, we selected \(\rho\) adaptively \cite{Huang2016}: 
\begin{align}
    \rho_{\Bk} &= \frac{\fnorm{\A \Dk}^2}{R}, \qquad
    \rho_\A = \sum_{k=1}^k \frac{\fnorm{\Bk \Dk}^2}{R}, \\
    \rho_{\Dk} &= \frac{1}{R}\Trace{\A \Tra \A * \Bk\Tra\Bk} \nonumber.
\end{align}

\section{Experiments}
For all models, we used our Python implementations, linked in the paper repository on GitHub\footnote{\url{https://github.com/MarieRoald/PARAFAC2-AOADMM-EUSIPCO21}}. The flexible coupling PARAFAC2 with hierarchical non-negative least squares algorithm (HALS) was implemented closely following the MATLAB implementation by Cohen and Bro \cite{CoBr18}. The implementation use the same hierarchical non-negative least squares algorithm \cite{GiGl12}, default parameter values and initialisation scheme. For the proximal operator of the total variation (TV) seminorm, we used the publicly available C implementation \cite{Co17} of the improved direct TV denoising algorithm presented in \cite{Co13}. We set both the relative and absolute tolerance equal to \(10^{-5}\) for the inner loops (the ADMM subproblems) and \(10^{-10}\) for the outer loop (the overall fitting procedure).

To measure convergence we used the relative SSE:
\begin{equation}
    \text{Rel. SSE} = \frac{1}{\fnorm{\X}^2} \sum_{k=1}^K \fnorm{\A \Dk \Bk \Tra - \Xk}^2.
\end{equation}
Also, we measured if the different models recovered the true components with the factor match score (FMS), given by:
\begin{equation}
    \text{FMS} = \frac{1}{R} \sum_{r=1}^R \MC{A}{r}\Tra \MC{\hat{A}}{r} \MC{B}{r}\Tra \MC{\hat{B}}{r} \MC{C}{r}\Tra \MC{\hat{C}}{r},
\end{equation}
where the hat represents the estimated component after solving the permutation indeterminacy. The \( \MC{B}{r} \) and \(\MC{c}{r}\)-vectors contain the concatenations of the \(r\)-th column of all \(\Bk\) matrices and the \(r\)-th diagonal entry of all \(\Dk\) matrices, respectively. All component vectors are normalised before computing the FMS.

To evaluate the AO-ADMM approach, we use a simulation setup inspired by \cite{CoBr18}. The elements of \(\A\) and \(\Dk\) factor matrices were respectively drawn from a truncated normal distribution and a uniform distribution between 0.1 and 1.1 (to avoid near zero elements in the \(\Dk\) matrices, which can hinder recovery of the \(\Bk\) matrices \cite{KiTeBr99}). The \(\Bk\) factor matrices were obtained by first generating a “blueprint matrix”, \(\hat{\M{B}}\) tailored to the constraint we wished to impose. The rows of \(\hat{\M{B}}\)  were subsequently cyclically shifted to obtain \(\Bk\) matrices, setting \(\left[\Bk\right]_{j r} = \hat{\M{B}}_{j_k r}\), with \(j_k = ((j + k) \mod J)\).

For each experiment, we created 50 random datasets. We constructed tensor slices, \(\Xk\), based on \eqref{eq:pf2} using known factor matrices. We let \(\T{X}\) be the tensor with frontal slices given by our data matrices, and added random noise according to
\begin{equation}
    \T{X}_{\text{noisy}} = \T{X} + \eta\fnorm{\T{X}}\tfrac{\T{E}}{\fnorm{\T{E}}},    
\end{equation}
where \(\eta\) is the noise level and \(\T{E}_{ijk} \sim \mathcal{N}(0, 1)\).

For each dataset, we fit models with five random initialisations, and kept the components that achieved lowest  final cost value. Non-negativity constraints were always imposed on the \(\Dk\)-matrices, to resolve the sign-indeterminacy of the PARAFAC2 model \cite{tensor:parafac2:Ha72,He13}. 

\subsection{Non-negativity constraints}
\begin{figure}
    \centering
    \begin{subfigure}[b]{0.32\linewidth}
    \includegraphics{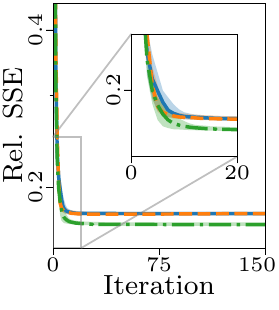}
    \end{subfigure}
    \begin{subfigure}[b]{0.32\linewidth}
    \includegraphics{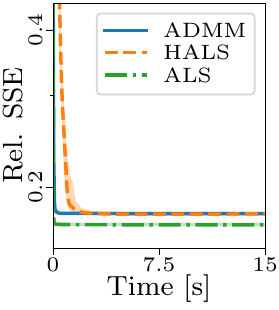}
    \end{subfigure}
    \begin{subfigure}[b]{0.32\linewidth}
    \includegraphics{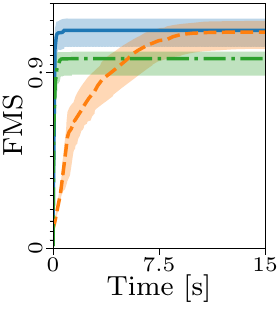}
    \end{subfigure}
    \caption{Diagnostic plots for the different datasets. The lines show the median values.}
    \label{fig:nn.results}
\end{figure}

To assess the performance of the AO-ADMM based algorithm for fitting a PARAFAC2 model with non-negativity constraints, we compared speed and accuracy with both HALS and the standard unregularised ALS. We generated the \(\hat{\M{B}}\) matrices with elements drawn from a truncated normal distribution (setup 1). The noise levels were set to 0.33 and 0.5. For AO-ADMM and HALS, non-negativity was imposed on all components, whereas for ALS, non-negativity was only imposed on the \(\A\) and \(\ForAllK{\Dk}\) matrices. The diagnostic plots for \(\eta = 0.5\) are shown in \cref{fig:nn.results}. Diagnostic plots demonstrate that both non-negative PARAFAC2 algorithms outperform ALS with respect to FMS. Moreover, the AO-ADMM algorithm is as fast as the traditional ALS algorithm and orders of magnitude faster than the flexible coupling approach. We observed the same behaviour for \(\eta = 0.33\) (see supplementary material).
\subsection{Structure imposing regularisation}
The AO-ADMM algorithm also allows for structure imposing regularisation such as graph Laplacian regularisation (\(\gamma_L \text{Tr}(\Bk\Tra \M{L} \Bk)\)) \cite{SmKo03} and total variation regularisation (\(\gamma_{TV}\|\Bk\|_{\text{TV}}\)). To assess the effectiveness of graph Laplacian regularisation, we set the components of \(\hat{\B}\) equal to emission spectra from a fluorescence spectroscopy dataset \cite{Br97} (setup 2). These spectra are smooth, i.e. neighbouring wavelengths have similar values, which makes graph Laplacian regularisation sensible. To impose smoothness, we set the graph Laplacian penalty function to \({\small \reg{\B}{\Bk} = \gamma_L \sum_{j r} \left(\left[\B_k\right]_{j r} - \left[\B_k\right]_{j+1 r}\right)^2}\). 

For assessing total variation regularisation, we used piecewise constant functions with 6 jumps whose derivatives summed to zero as the components of \(\hat{\B}\) (setup 3). For both structure imposing regularisation experiments, we tested with two different noise levels: \(\eta \in \{0.33, 0.5\}\) and we imposed ridge regularisation on \(\A\) and \(\Dk\) (\(\gamma_r (\fnorm{\M{A}} + \sum_k \fnorm{\Dk} ) \)). The regularisation parameters were found through a grid search (details in supplementary material). We also fitted PARAFAC2 models to the same datasets with the traditional unregularised ALS algorithm \cite{KiTeBr99} as a baseline.

\begin{figure}
    \centering
    \begin{subfigure}[b]{\linewidth}
        \includegraphics{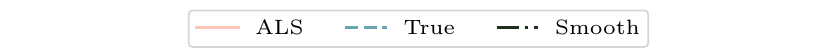}
    \end{subfigure}
    \begin{subfigure}[b]{\linewidth}
        \includegraphics{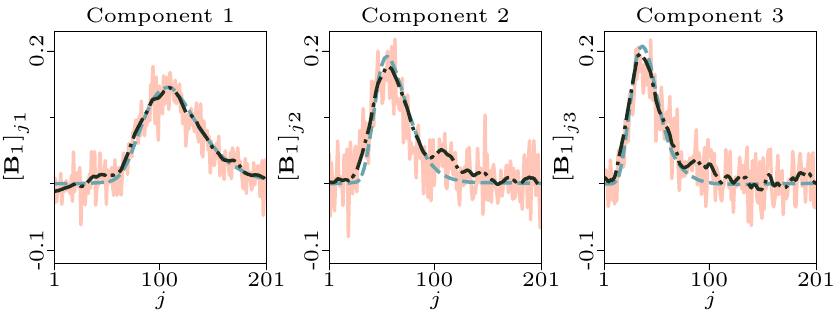}
    \end{subfigure}
    \caption{The true and estimated columns of \({\small\Mn{B}{1}}\) for one dataset with \({\small\eta\hspace{-0.2em}=\hspace{-0.2em}0.5}\). The smoothness regularised components were fitted with \({\small\gamma_r \hspace{-0.2em}=\hspace{-0.2em} 0.01}\) and \({\small\gamma_{L}\hspace{-0.2em}=\hspace{-0.2em}1000}\).  \({\small\Mn{X}{1}}\) had a signal to noise ratio of 0.1~dB.}
    \label{fig:smooth.results}
\end{figure}

The structure imposing regularisation helped recovery for both setup 2 and 3. For most parameter combinations, we observed an increase in FMS compared to unregularised models. The performance degraded only with a very high degree of regularisation. \cref{tab:results} shows the results for the parameters that obtained the highest mean FMS. In \cref{fig:smooth.results} we see that the graph Laplacian regularised models led to smooth components and \cref{fig:tv.results} shows that the TV regularisation  produced piecewise constant components (see supplementary for animated plots). For both setups, the ALS algorithm yielded noisy components.

\begin{figure}
    \centering
    \begin{subfigure}[b]{\linewidth}
        \includegraphics{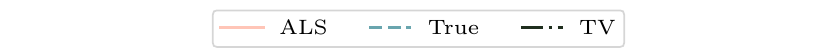}
    \end{subfigure}
    \begin{subfigure}[b]{\linewidth}
        \includegraphics{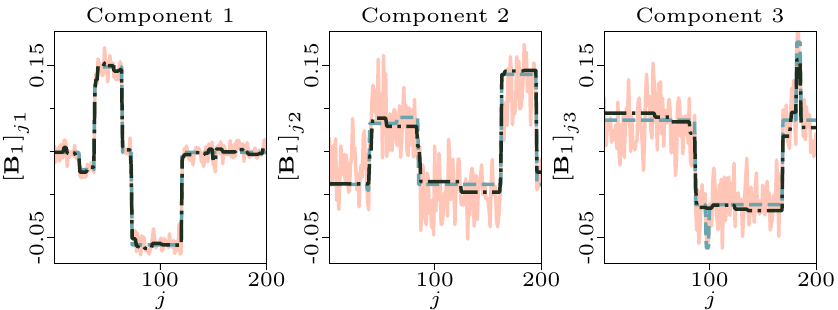}
    \end{subfigure}
    \caption{The true and estimated columns of \({\small\Mn{B}{1}}\) for one dataset with \({\small \eta\hspace{-0.2em}=\hspace{-0.2em}0.5}\). The TV regularised components were fitted with \({\small\gamma_r \hspace{-0.2em}= \hspace{-0.2em}10}\) and \({\small\gamma_{TV}\hspace{-0.2em}=\hspace{-0.2em}10}\). \({\small\Mn{X}{1}}\) had a signal to noise ratio of 5.9~dB.}
    \label{fig:tv.results}
\end{figure}

\begin{table}[]
    \centering
    \caption{FMS results from experiments with structure imposing regularisation.}
    \label{tab:results}
    \begin{tabular}{@{}rr@{\hspace{0.5em}}rrr@{}}
    \toprule
     & \multicolumn{2}{c}{Setup 2} &  \multicolumn{2}{c}{Setup 3}\\
        \cmidrule(lr){2-3}  \cmidrule(l){4-5} 
        \textbf{Method} & \(\eta=0.33\) & \(\eta=0.5\) & \(\eta=0.33\) & \(\eta=0.5\)\\
        \cmidrule(r){1-1} \cmidrule(lr){2-2} \cmidrule(lr){3-3} \cmidrule(lr){4-4} \cmidrule(l){5-5} 
        AO-ADMM & \(0.99 \pm 0.01\) & \(0.98 \pm 0.01\) & \(0.98 \pm 0.02\) & \(0.96 \pm 0.05\)\\
        ALS & \(0.96 \pm 0.01\) & \(0.92 \pm 0.01\) & \(0.92 \pm 0.06\) & \(0.86 \pm 0.08\)\\
    \bottomrule
    \end{tabular}
\end{table}

\section{Conclusion}
In this work, we proposed an efficient AO-ADMM-based algorithmic framework for fitting PARAFAC2 models with regularisation. Using the proposed approach, we can fit PARAFAC2 models with any proximable regularisation penalty on all factor matrices of the decomposition. Our experiments on simulated data demonstrate that the AO-ADMM framework is faster than the flexible coupling approach for non-negative PARAFAC2. Moreover, we show that our approach can successfully apply structure imposing regularisation, such as TV and graph Laplacian regularisation, on the evolving mode of a PARAFAC2 model.
\printbibliography

@Article{AcBiBiBr07,
  author =        {Evrim Acar and Canan A. Bingol and Haluk Bingol and Rasmus Bro and Bulent Yener},
  title =         {Multiway Analysis of Epilepsy Tensors},
  journal =       {Bioinformatics},
  year =          {2007},
  volume =        {23},
  number =        {13},
  pages =         {i10-i18},
  doi =           {10.1093/bioinformatics/btm210}
}

@inproceedings{YiAfHoChZhSu20,
  title={{LogPar}: Logistic {PARAFAC2} Factorization for Temporal Binary Data with Missing Values},
  author={Yin, Kejing and Afshar, Ardavan and Ho, Joyce C and Cheung, William K and Zhang, Chao and Sun, Jimeng},
  booktitle={Proc. 26th ACM SIGKDD Int. Conf. Knowl. Discov. and Data Mining},
  pages={1625--1635},
  year={2020}
}

@incollection{SmKo03,
  title={Kernels and regularization on graphs},
  author={Smola, Alexander J and Kondor, Risi},
  booktitle={Learning theory and kernel machines},
  pages={144--158},
  year={2003},
  publisher={Springer}
}

@online{Co17,
    year={2019},
    title={Software},
    author={Condat, Laurent},
    url={https://lcondat.github.io/software.html},
    note={[Accessed: 2020-10-19]}
}

@article{Co13,
  title={A direct algorithm for 1-{D} total variation denoising},
  author={L. Condat},
  journal={IEEE Signal Process. Letters},
  volume={20},
  number={11},
  pages={1054--1057},
  year={2013},
  publisher={IEEE}
}

@inproceedings{ScCoAc20,
  title={An Optimization Framework for Regularized Linearly Coupled Matrix-Tensor Factorization},
  author={Schenker, Carla and Cohen, Jeremy E and Acar, Evrim},
  booktitle={Proc. 28th Eur. Signal Process. Conf. (EUSIPCO)},
  pages={985--989},
  year={2020},
  organization={IEEE}
}

@article{He13,
  title={The special sign indeterminacy of the direct-fitting {P}arafac2 model: Some implications, cautions, and recommendations for Simultaneous Component Analysis},
  author={N. E. Helwig},
  journal={Psychometrika},
  volume={78},
  number={4},
  pages={725--739},
  year={2013},
  publisher={Springer}
}

@article{tensordec:KoTaBa09,
  title={Tensor decompositions and applications},
  author={T. G. Kolda and B. W. Bader},
  journal={SIAM Rev.},
  volume={51},
  number={3},
  pages={455--500},
  year={2009},
  publisher={SIAM}
}

@Article{tensor:parafac2:Ha72,
  author  = {R. A. Harshman},
  journal = {UCLA working papers in phonetics},
  title   = {{PARAFAC2}: {M}athematical and technical notes},
  year    = {1972},
  pages   = {30--44},
  volume  = {22},
}

@InProceedings{CoBr18,
  author    = {J. E. Cohen and R. Bro},
  booktitle = {LVA/ICA'18},
  title     = {Nonnegative {PARAFAC2}: A Flexible Coupling Approach},
  year      = {2018},
  pages     = {89--98},
}

@Article{PARAFAC:Ha70,
  author  = {R. A. Harshman},
  journal = {UCLA working papers in phonetics},
  title   = {Foundations of the {PARAFAC} procedure: Models and conditions for an ``explanatory'' multi-modal factor analysis},
  year    = {1970},
  pages   = {1-84},
  volume  = {16},
}

@Article{PARAFAC:Ca70,
  author  = {J. D. Carroll and J. J. Chang},
  journal = {Psychometrika},
  title   = {Analysis of individual differences in multidimensional scaling via an {N}-way generalization of ``{Eckart-Young}'' decomposition},
  year    = {1970},
  issn    = {1860-0980},
  number  = {3},
  pages   = {283--319},
  volume  = {35},
  doi     = {10.1007/BF02310791},
  url     = {https://doi.org/10.1007/BF02310791},
}

@InProceedings{tensor:smoothness:parafac2:COPA:AfPePaSeHoSe18,
  author    = {A. Afshar and I. Perros and E. E. Papalexakis and E. Searles and J. Ho and J. Sun},
  booktitle = {ACM Int. Conf. on Inf. and Knowl. Management},
  title     = {{COPA}: {C}onstrained {PARAFAC2} for {S}parse \& {L}arge {D}atasets},
  year      = {2018},
  pages     = {793--802},
  numpages  = {10},
}

@Article{parafac2:BrAnKi99,
  author   = {R. Bro and C. A. Andersson and H. A. L. Kiers},
  journal  = {J. Chemom.},
  title    = {{PARAFAC2 - Part II. Modeling chromatographic data with retention time shifts}},
  year     = {1999},
  issn     = {0886-9383},
  number   = {3-4},
  pages    = {295--309},
  volume   = {13},
  doi      = {10.1002/(SICI)1099-128X(199905/08)13:3/4<295::AID-CEM547>3.0.CO;2-Y},
}

@Article{Huang2016,
  author    = {K. Huang and N. D. Sidiropoulos and A. P. Liavas},
  journal   = {IEEE Trans. Signal Process.},
  title     = {A flexible and efficient algorithmic framework for constrained matrix and tensor factorization},
  year      = {2016},
  number    = {19},
  pages     = {5052--5065},
  volume    = {64},
  publisher = {IEEE},
}

@article{PaBo14,
  title={Proximal algorithms},
  author={Parikh, Neal and Boyd, Stephen},
  journal={Found. Trends Mach. Learn.},
  volume={1},
  number={3},
  pages={127--239},
  year={2014},
  publisher={Now Publishers Inc. Hanover, MA, USA}
}

@Article{BoPaChPeEc11,
  author  = {S. Boyd and N. Parikh and E. Chu and B. Peleato and J. Eckstein},
  journal = {Found. Trends Mach. Learn.},
  title   = {Distributed optimization and statistical learning via the alternating direction method of multipliers},
  year    = {2011},
  month   = jan,
  number  = {1},
  pages   = {1--122},
  volume  = {3},
}

@Article{KiTeBr99,
  author  = {H. A. L. Kiers and J. M. F. {Ten Berge} and R. Bro},
  journal = {J. Chemom.},
  title   = {{PARAFAC2} - {Part I. A} direct fitting algorithm for the {PARAFAC2} model},
  year    = {1999},
  number  = {3-4},
  pages   = {275--294},
  volume  = {13},
}

@Article{tensor:smoothness:parafac2:He17,
  author   = {N. E. Helwig},
  journal  = {Biom. J.},
  title    = {Estimating latent trends in multivariate longitudinal data via {Parafac2} with functional and structural constraints},
  year     = {2017},
  number   = {4},
  pages    = {783-803},
  volume   = {59},
  doi      = {10.1002/bimj.201600045},
  eprint   = {https://onlinelibrary.wiley.com/doi/pdf/10.1002/bimj.201600045},
  url      = {https://onlinelibrary.wiley.com/doi/abs/10.1002/bimj.201600045},
}

@InProceedings{RoShJi19,
  author      = {M. Roald and S. Bhinge and C. Jia and V. Calhoun and T. Adali and E. Acar},
  title       = {Tracing Network Evolution using the {PARAFAC2} model},
  booktitle   = {Proc. Int. Conf. on Acoust., Speech, and Signal Process.},
  year        = {2020},
  doi         = {10.1109/ICASSP40776.2020.9053902},
  institution = {arXiv:1911.02926v1},
}

@InProceedings{ChBaBr07,
  author    = {P. A. Chew and B. W. Bader and T. G. Kolda and A. Abdelali},
  booktitle = {Proc. 13th ACM SIGKDD Int. Conf Knowl Discov and Data Mining},
  title     = {Cross-Language Information Retrieval Using {PARAFAC2}},
  year      = {2007},
  pages     = {143–152},
  doi       = {10.1145/1281192.1281211},
  numpages  = {10},
}

@article{MoHaHePaAr06,
  title={Parallel factor analysis as an exploratory tool for wavelet transformed event-related {EEG}},
  author={M{\o}rup, Morten and Hansen, Lars Kai and Herrmann, Christoph S and Parnas, Josef and Arnfred, Sidse M},
  journal={NeuroImage},
  volume={29},
  number={3},
  pages={938--947},
  year={2006},
  publisher={Elsevier}
}

@article{Br97,
  title={{PARAFAC}. Tutorial and applications},
  author={Bro, Rasmus},
  journal={Chemom. and Intel. Lab. Systems},
  volume={38},
  number={2},
  pages={149--172},
  year={1997},
  publisher={Amsterdam; New York: Elsevier Science Pub. Co., 1986-}
}

@article{GiGl12,
  title={Accelerated multiplicative updates and hierarchical {ALS} algorithms for nonnegative matrix factorization},
  author={N. Gillis and F. Glineur},
  journal={Neural Comput.},
  volume={24},
  number={4},
  pages={1085--1105},
  year={2012},
  publisher={MIT Press}
}

\end{document}